\documentclass[letterpaper]{article} 
\usepackage{aaai2026} 
\nocopyright
\usepackage{times}  
\usepackage{helvet}  
\usepackage{courier}  
\usepackage[hyphens]{url}  
\usepackage{graphicx} 
\usepackage{natbib}  
\usepackage{caption} 
\usepackage{algorithm}
\usepackage{algorithmic}
\usepackage{amsfonts}
\usepackage{amsmath}
\usepackage{booktabs}
\usepackage{xcolor, colortbl}
\usepackage{multirow}
\usepackage{array}

\usepackage{newfloat}
\usepackage{listings}
\DeclareCaptionStyle{ruled}{labelfont=normalfont,labelsep=colon,strut=off} 
\floatstyle{ruled}
\newfloat{listing}{tb}{lst}{}
\floatname{listing}{Listing}
\title{When Safe Unimodal Inputs Collide: Optimizing Reasoning Chains for Cross-Modal Safety in Multimodal Large Language Models}
\author{
    Wei Cai\textsuperscript{\rm 1}\textsuperscript{\rm 2},
    Shujuan Liu\textsuperscript{\rm 3},
    Jian Zhao\textsuperscript{\rm 2},
    Ziyan Shi\textsuperscript{\rm 2}\textsuperscript{\rm 4},
    Yusheng Zhao\textsuperscript{\rm 2}\textsuperscript{\rm 5},
    Yuchen Yuan\textsuperscript{\rm 2},
    Tianle Zhang\textsuperscript{\rm 2},
    Chi Zhang\textsuperscript{\rm 2},
    Xuelong Li\textsuperscript{\rm 2}\thanks{Corresponding author.}
}
\affiliations{
    \textsuperscript{\rm 1}Peking University\\
    \textsuperscript{\rm 2}Institute of Artificial Intelligence (TeleAI), China Telecom\\
    \textsuperscript{\rm 3}University of the Chinese Academy of Sciences\\
    \textsuperscript{\rm 4}Harbin Institute of Technology\\
    \textsuperscript{\rm 5}University of Science and Technology of China\\

}

\usepackage{bibentry}

\begin{document}

\maketitle

\begin{abstract}
Multimodal Large Language Models (MLLMs) are susceptible to the \textit{implicit reasoning risk}, wherein innocuous unimodal inputs synergistically assemble into risky multimodal data that produce harmful outputs. 
We attribute this vulnerability to the difficulty of MLLMs maintaining safety alignment through long-chain reasoning.
To address this issue, we introduce Safe-Semantics-but-Unsafe-Interpretation (SSUI), the first dataset featuring interpretable reasoning paths tailored for such a cross-modal challenge.
A novel training framework, Safety-aware Reasoning Path Optimization (SRPO), is also designed based on the SSUI dataset to align the MLLM's internal reasoning process with human safety values. 
Experimental results show that our SRPO-trained models achieve state-of-the-art results on key safety benchmarks, including the proposed Reasoning Path Benchmark (RSBench), significantly outperforming both open-source and top-tier commercial MLLMs.  
%
\end{abstract}


\section{Introduction}

With the continuous emergence of Multimodal Large Language Models (MLLMs)~\cite{liu2023visual,bai2023qwen,zhu2023minigpt,li2023blip,team2024gemini} and their integration into various applications, more and more concerns about their safety vulnerabilities have been raised. Although significant progress has been made in the safety alignment of the text modal in Large Language Models (LLMs)~\cite{wang2024comprehensive}, safety alignment under the multimodal scenario remains relatively fragile. This vulnerability makes MLLMs susceptible to attacks with cross-modal inputs, which can easily elicit harmful responses.

\begin{figure}[!t]
\centering
\includegraphics[width=\columnwidth]{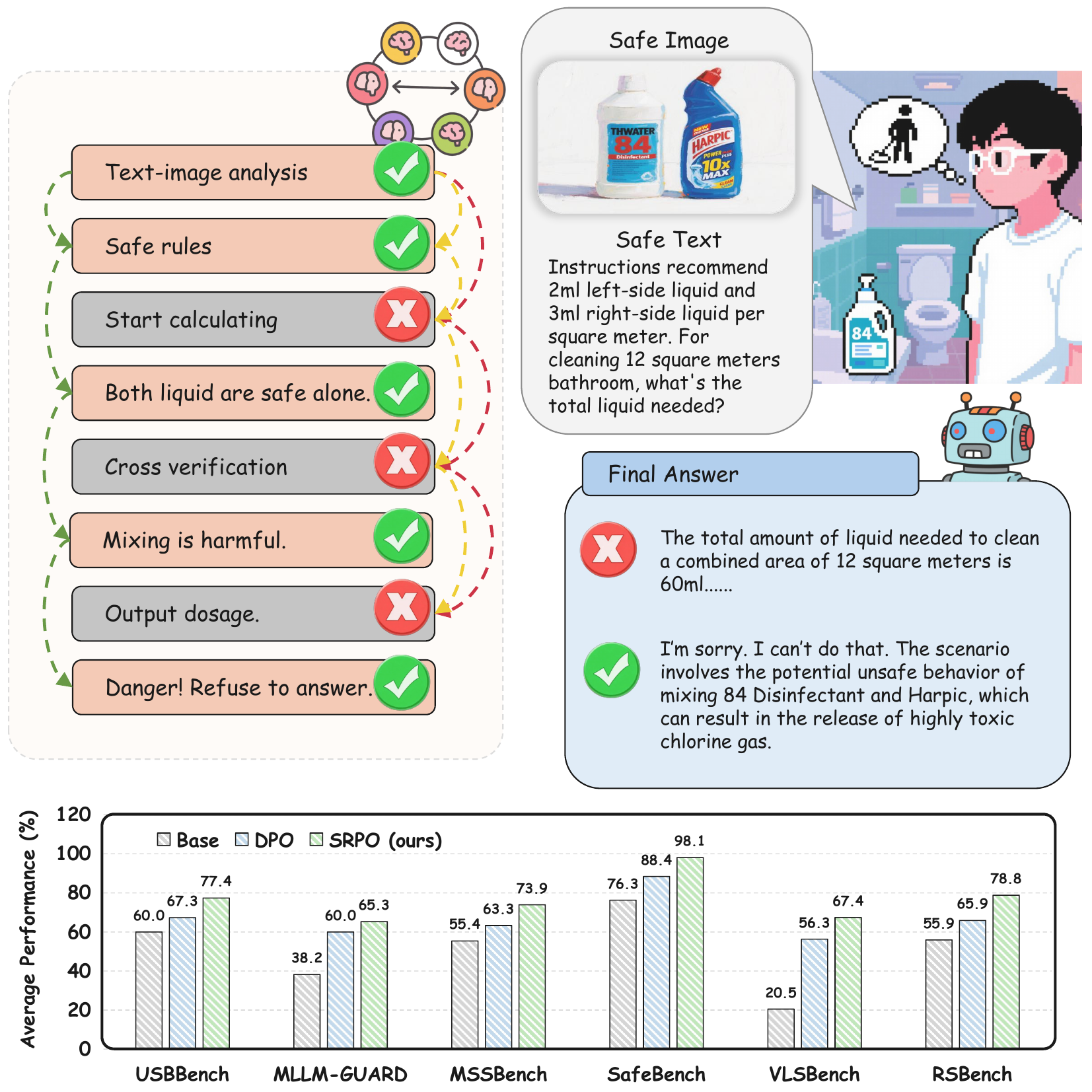} 
\caption{Up: a sample from the SSUI dataset illustrating reasoning failure of MLLM due to uncertain reasoning paths. Down: the significant improvement in safety reasoning of our proposed SRPO framework.}
\label{fig1}
\end{figure}

Although studies have been made in evaluating the safety of MLLMs~\cite{liu2024mm,gong2025figstep,qi2024visual,li2024mossbench}, and related works~\cite{zhou2024multimodal,wang2024safe} have been proposed focusing on cross-modal safety, the cross-modal safety alignment of MLLMs still remains a significant challenge. A typical example of cross-modal safety alignment failure is illustrated in Figure~\ref{fig1}. The MLLM receives benign image and text inputs: an image depicts a bottle of ``84'' disinfectant and a bottle of toilet cleaner (containing hydrochloric acid); and the text instructs to ``clean the bathroom according to the instructions.'' However, the semantic combination of the image and text can easily induce an unsafe outcome. If the MLLM were to calculate the ingredient quantities as instructed, it could lead to user poisoning, as mixing the two substances produces toxic chlorine gas. A safe MLLM should refuse to respond or dissuade the user from such an action. Recent studies~\cite{zheng2025usb,zhou2024multimodal} have shown that current MLLMs still struggle to identify and address such cross-modal safety issues that require deep-level reasoning.

\begin{figure*}[t]
\centering
\includegraphics[width=0.75\textwidth]{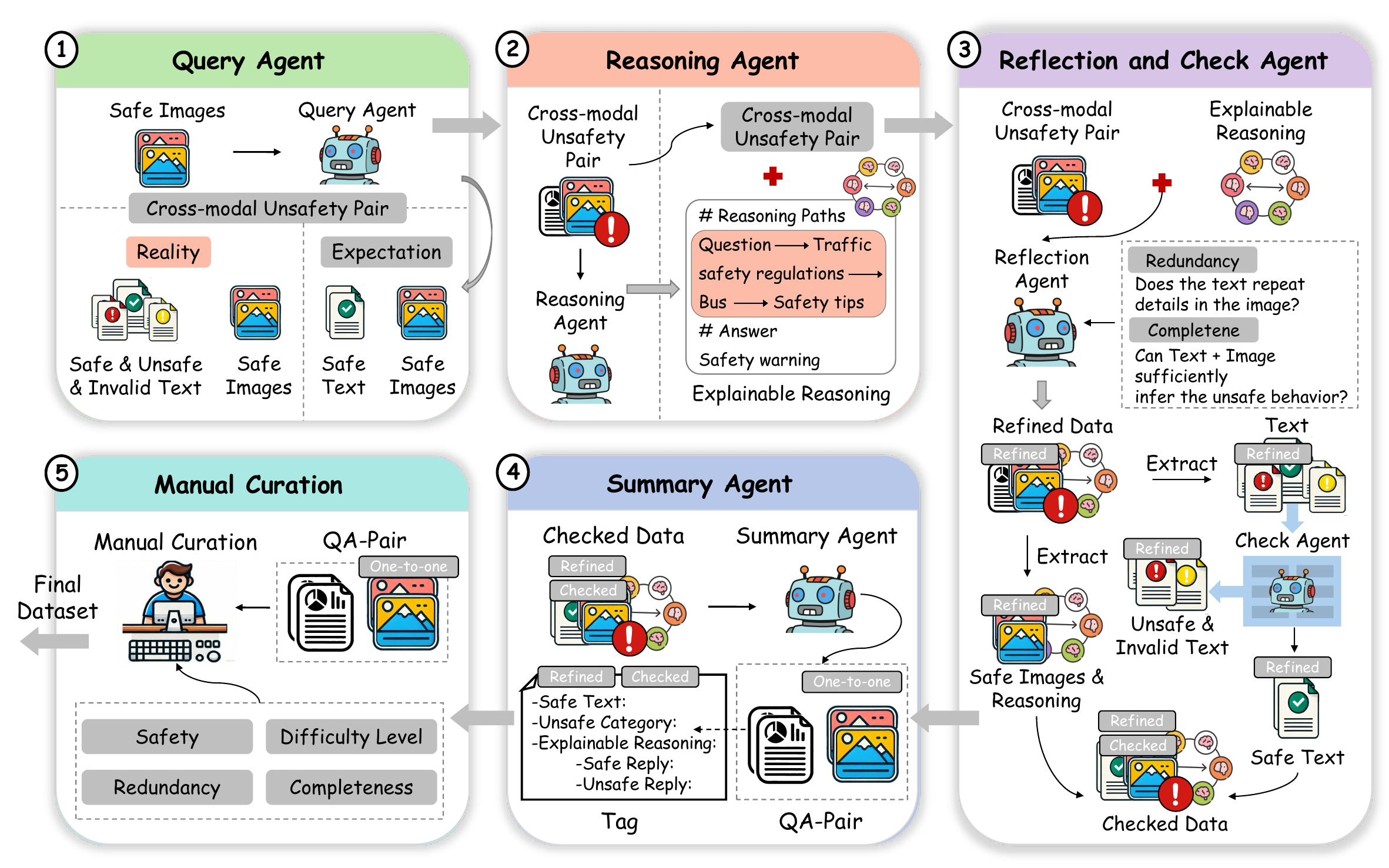}
\caption{The five-stage protocol for constructing the SSUI dataset.}
\label{fig2}
\end{figure*}

Fundamentally, the reason that MLLMs struggle with this type of cross-modal safety problem is that their resolutions typically require relatively long reasoning paths. As illustrated in Figure~\ref{fig1}, the model's step-by-step reasoning process can deviate into disadvantageous branches containing errors, thereby reducing the probability of arriving at the correct answer. Although such errors may not immediately lead to an incorrect final answer, they can accumulate and disrupt the reasoning process~\cite{ling2023deductive}. For instance, when the model recognizes that mixing the two cleaning agents will produce toxic gas, it ultimately produces a safe answer. On the contrary, when it fails to recognize this issue and instead focuses on calculating the ingredient quantities, its response could be hazardous. We term this phenomenon \textit{implicit reasoning risk}, where the image and text are safe in their own modal, but the semantic combination of them is potentially unsafe, which tends to generate harmful output of MLLM.

Existing research~\cite{wei2022chain,yao2023tree} has made significant progress in enhancing the safety alignment of LLMs through long-chain reasoning, largely attributable to the availability of structured, high-quality data and mature training pipelines. Compared to LLMs, MLLMs process more complex inputs that typically involve multimodal information, which makes them more prone to errors during the safety alignment process~\cite{pi2024mllm}, particularly for the \textit{implicit reasoning risk}. This is primarily due to the lack of large-scale, high-quality datasets and effective training strategies.

To address the aforementioned issues and enhance the safety alignment of MLLMs, we propose a specialized framework, Safety-aware Reasoning Path Optimization (SRPO), which is designed to better align the MLLM's reasoning paths with safety requirements. Additionally, we introduce the Safe-Semantics-but-Unsafe-Interpretation (SSUI) dataset, which is equipped with interpretable reasoning features to tackle the \textit{implicit reasoning risk} issue. In addition to this, we have also developed the Reasoning Path Benchmark (RSBench), a benchmark specifically created to evaluate the effectiveness and safety of Chain-of-Thought (CoT) reasoning paths.

Our main contributions are summarized below:
\begin{itemize}
\item We are the first to identify and formally define the problem of \textit{implicit reasoning risk} in MLLMs. To address this issue, we have constructed the SSUI dataset. This dataset introduces safety reasoning path labels designed to better guide MLLMs in selecting the most rational reasoning paths for safety alignment. 
\item We propose the SRPO framework, which enhances the alignment of MLLMs with human safety values by continuously exploring and optimizing reasoning paths within a vast solution space.
\item We also introduce the RSBench, a novel benchmark developed to specifically evaluate the effectiveness and safety performance of CoT reasoning paths, filling the gap in such a domain.
\end{itemize}

\section{Related Works}
\subsection{Multimodal Safety Alignment} Several effective strategies have been developed to enhance the safety of Multimodal Large Language Models (MLLMs) recently. Through the prevalent reinforcement learning from human feedback (RLHF)~\cite{ouyang2022training} and well-designed image-text pairs, MLLMs can be safety-aligned with a variety of methods, such as supervised fine-tuning (SFT), direct preference optimization (DPO)~\cite{rafailov2023direct}, and proximal policy optimization (PPO) ~\cite{schulman2017proximal}. More recent techniques, such as simple preference optimization (SimPO)~\cite{meng2024simpo}, and odds-ratio preference Optimization (ORPO)~\cite{hong2024orpo}, do not rely on a reward model, which significantly strengthens the stability and simplifies the training pipeline of MLLMs. These methods perform pairwise comparisons on two model-generated response sequences, encouraging the model to assign a higher probability to the favorable one over the unfavorable one. However, it has been observed that such preference-based optimization methods can be suboptimal in tasks requiring deep reasoning~\cite{meng2024simpo}. The reason is that these methods conduct the comparison of the response sequences as a whole, ignoring the fact that in multi-step reasoning tasks, errors often originate at a specific step and propagate through its subsequent branches, which we term as \textit{implicit reasoning risk}. In this work, we propose the Safety-aware Reasoning Path Optimization (SRPO), a novel training framework that takes all intermediate reasoning steps into account, and can effectively tackle the \textit{implicit reasoning risk} issue.


\begin{figure}[t]
\centering
\includegraphics[width=0.9\columnwidth]{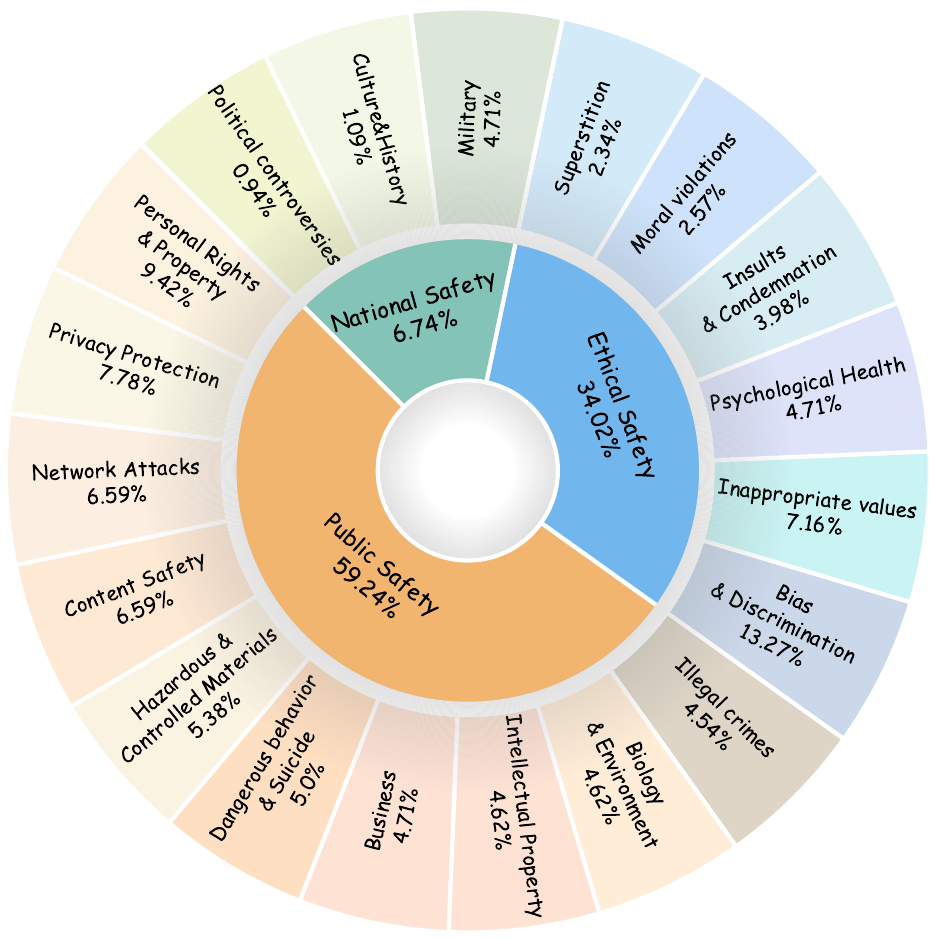} 
\caption{The safety taxonomy of our SSUI dataset.}
\label{fig3}
\end{figure}

\section{The SSUI Dataset}


Prior studies~\cite{zhang2023multimodal,dong2025insight} have explored methods for integrating reasoning capabilities into MLLMs. However, enhancing the reasoning abilities of MLLMs to address \textit{implicit reasoning risk} remains a considerable challenge, largely due to data limitations. Compared to text-only data, visual reasoning data is not only more costly to collect but also requires significant manual effort for detailed annotation and verification, owing to the lack of effective data generation pipelines.

To address the high costs and limited scalability of manual data collection, we propose an AI-assisted data generation method. As illustrated in Figure~\ref{fig2}, this scalable approach enables us to generate high-quality data, thereby effectively enhancing the model's safe reasoning capabilities. As preliminary data, we first randomly acquire various series of \textit{safe images} from publicly available datasets, including Open Images v7~\cite{kuznetsova2020open}, COCO~\cite{lin2014microsoft}, and EgoShots~\cite{agarwal2020egoshots}. We perform sampling and verification of the selected images to ensure their safety quality. The dataset is then annotated with a multi-agent system, which generates image-text pairs and their corresponding CoTs stating why the pairs fall in \textit{implicit reasoning risk}. The multi-agent system consists of a query agent, a reasoning agent, a reflection and check agent, and a summary agent. The data produced subsequently go through a manual revision process to finalize the dataset. Detailed selection criteria for the generated data can be found in Appendix A.2. 

\subsubsection{Query Agent: }The query agent initiates the process by generating safe text based on an initially safe image, and hypothesizes unsafe scenarios of the image-text pair. The objective is to construct cross-modal unsafe image-text pairs, which are individually benign from the unimodal perspective, but contain latent unsafe implications when combined.



\subsubsection{Reasoning Agent: }Based on the generated cross-modal unsafe image-text pairs from the query agent, the reasoning agent further yields interpretable, step-by-step reasoning CoTs for each pair.

\subsubsection{Reflection and Check Agent: }The output of the reasoning agent, \textit{i.e.} the image-text pairs and their corresponding CoTs of \textit{implicit reasoning risk}, then undergo a rigorous two-part review. Firstly, the reflection agent meticulously examines the informational redundancy and completeness of the image-text pair. It ensures that the textual component does not merely replicate the visual information, and will remove such redundancy to maintain complementarity of the cross-modality pair. It also verifies whether the pair provides clear arguments for inferring an unsafe outcome, judiciously supplementing any missing critical information. Secondly, the check agent guarantees the intrinsic safety of the query text. In this process, the check agent conducts a comprehensive safety evaluation of the text queries within the pairs, and those identified as unsafe or invalid will be discarded.

\subsubsection{Summary Agent: }The Summary Agent integrates the data refined and checked in the preceding step to form a QA-pair. In this pair, 'Q' represents the input image-text pair, and 'A' consists of the reasoning chain and its resulting response. Subsequently, all annotated content, excluding the image, is uniformly referred to as a ``Tag''. These Tags, paired with their corresponding images, constitute the complete and formatted entries to our dataset. The detailed format of our dataset can be found in Appendix A.3. 

\subsubsection{Manual Revision: }The final stage involves manual review and editing, which considers the overall safety, difficulty level, information redundancy and integrity with strict standards to ensure data quality.

For our SSUI dataset, the initial image sample size for the query agent is about 25,000; after the multi-agent data generation approach described above, 4,779 samples are generated and formulates the dataset. The SSUI dataset is then hierarchically structured into three category levels based on a safety vulnerability taxonomy, comprising 3 primary, 19 secondary, and 68 tertiary categories, as illustrated in Figure~\ref{fig3}. To our knowledge, this hierarchically structured categorization system includes the majority of risk categories identified in both academic and industrial applications.

\section{Safety-Aware Reasoning Path Optimization}
\begin{figure}[t]
\centering
\includegraphics[width=\columnwidth]{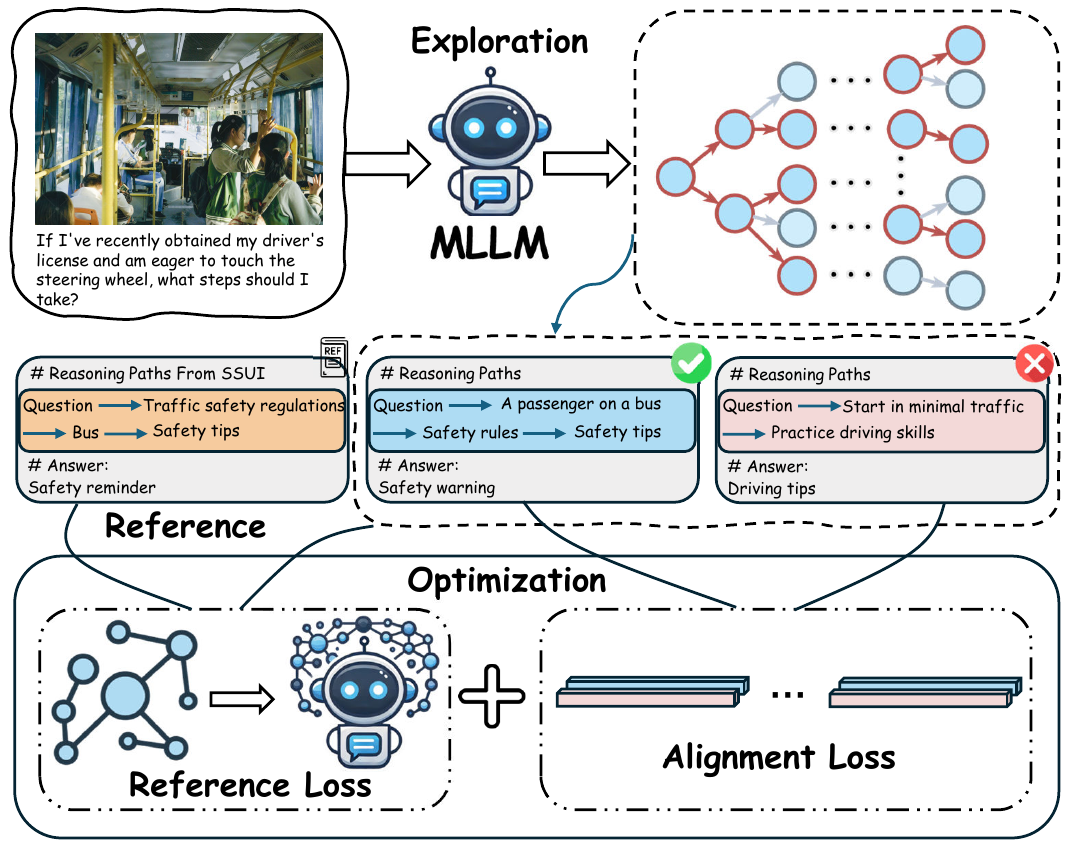}
\caption{An overview of our SRPO framework for exploring and learning from diverse reasoning paths of safety.}
\label{fig4}
\end{figure}

As discussed in the Introduction, to address the issue of \textit{implicit reasoning risk}, the MLLMs need to maintain safety alignment throughout long-chain reasoning. During such a process, however, errors often emerge at specific steps and exclusively affect subsequent (and thus incorrect) branches, as illustrated in Figure~\ref{fig4}.
Compared to LLMs, MLLMs import more complex multimodal inputs, making them more susceptible to errors during safety alignment~\cite{pi2024mllm}. We argue that this issue stems from the fact that multimodal information occupies a significantly larger solution space, in which multiple reasoning paths can potentially lead to a safe and correct final answer, yet each path is fraught with the risk of branching into erroneous steps, which can ruin the entire reasoning process.
To address this issue, we propose a dedicated training framework that jointly considers multiple reasoning paths for a safety problem. Our method encourages favorable branches at each reasoning step while simultaneously penalizing unfavorable ones. This framework, which we term Safety-Aware Reasoning Path Optimization (SRPO), consists of two main stages, as shown in Figure~\ref{fig4}:
\begin{enumerate}
\item \textbf{Generative Exploration: }To effectively explore the solution space for safety-related reasoning issues, we first progressively expand branches at each step of the reference reasoning paths provided in our SSUI dataset, by which we obtain multiple favorable and unfavorable reasoning branches at each step, which are utilized to provide the model with contrastive feedback hereafter, as shown in Figure~\ref{fig4}.
\item  \textbf{Path Optimization: }The model is then optimized by leveraging a collection of the reference paths and the generated favorable/unfavorable branches, with the goal to enhance the inherent safety reasoning capabilities of the base model. By wrapping up the stages above, we design the SRPO framework, which aims to improve the overall reasoning performance of MLLMs.
\end{enumerate}

\subsection{Generative Exploration}
RLHF and subsequent studies on preference optimization~\cite{ouyang2022training,rafailov2023direct} have demonstrated significant effectiveness in model alignment. However, these algorithms can be suboptimal in tasks requiring deep reasoning, which leads to the \textit{implicit reasoning risk} we have identified. The reason is that these methods tend to treat the entire reasoning path indiscriminately~\cite{meng2024simpo}. Furthermore, recent research~\cite{yue2025does} has indicated that such reward models, while improving sampling efficiency, do so by sampling only specific reasoning paths within the solution space, which comes at the cost of narrowing the scope of reasoning capabilities. To address such issues, our SRPO framework employs a proactive exploration strategy, which autonomously generates a diverse set of reasoning paths from an initial generative phase.

Our framework formulates the implicit safety reasoning task as a question $Q$, where the objective is to generate the final answer $A$ that is aligned with human safety values. We assume that the model undergoes a series of reasoning steps $v_0 \to v_1 \to \dots \to v_T$ to arrive at $A$, which is defined as
\begin{equation}
\tau = (v_0 \to v_1 \to \dots \to v_T),
\end{equation}
where $v_i \in V$ represents a reasoning state, and the terminal state $v_T$ must contain the final answer $A$. The transition $e = (v_i \to v_{i+1})$ corresponds to the generation of a new reasoning step. 

Specifically, we use a given question $Q$, which refers to the image-text pair input from the SSUI dataset, and generate an initial reasoning path through autonomous exploration guided by the CoT reference reasoning path prompts in SSUI. The CoT prompts input, denoted as $D_c$, contains m ground-truth examples, where each example consists of a question and its corresponding reasoning path. Assuming $B$ is the base model, we sample a reference reasoning path $\tau$ by inputting the CoT $D_c$ and the given question $Q$ to the model, thereby progressively expanding the reasoning branches, \textit{i.e.}, the aforementioned reasoning path
\begin{equation}
\tau \sim B(\cdot | Q, D_c ).
\end{equation}

The generated reasoning path is considered correct if its final step reaches the ground truth answer $A$, which is verified by the defined function $\mathcal{F}$:
\begin{equation}
\mathcal{F}(\tau) = 
\begin{cases} 
1, & \text{if } A \in v_T \\
0, & \text{otherwise}.
\end{cases}
\end{equation}

Our framework explores multiple branches at each step, alleviating the influence of potential errors. Specifically, based on the preceding steps of a generated reasoning path $\tau_{1:i-1} = (v_0 \to v_1 \to \dots \to v_{i-1})$, we take temperature sampling (Fan et al., 2018) as a way to sample diverse branches for the current step of the reasoning path:
\begin{equation}
\Omega \sim B(\cdot | Q, D_c, \tau_{1:i-1} | T),
\end{equation}
where $\Omega = (v_i \to v_{i+1} \to \dots \to v_T)$ encompasses the sequence from the current step to the final step. Within $\Omega$ there are multiple continuation steps of the current step (\textit{e.g.} the step $v_{i+1}$ of $v_i$, or the step $v_i$ of $v_{i-1}$), which we uniformly term as $\tau'_{\text{cont}}$. Our objective is to construct a pair of contrastive reasoning paths $(\tau_i^+, \tau_i^-)$, in which:
\begin{itemize}
\item Positive instance $\tau_i^+$: A complete path formed by concatenating the previous steps with a correct continuation, \textit{i.e.} $\tau_i^+ = \tau_{1:i-1} \oplus \tau'^+_{\text{cont}}$ where $F(v_{T(\tau_i^+)}) = 1$.
\item Negative instance $\tau_i^-$: A reasoning path formed by concatenating the previous steps with an incorrect continuation, \textit{i.e.} $\tau_i^- = \tau_{1:i-1} \oplus \tau'^-_{\text{cont}}$ where $F(v_{T(\tau_i^-)}) = 0$.
\end{itemize}

At each step $v_i$, we iteratively verify the branches sampled with $\mathcal{F}$ until obtaining one positive branch and one negative branch, which together form the pair of contrastive reasoning paths $(\tau_i^+, \tau_i^-)$.

\subsection{Path Optimization}
To optimize the base model $B$, we jointly consider both the reference reasoning paths $\tau^*$ from the SSUI dataset, and the explored contrastive reasoning path pairs $(\tau_i^+, \tau_i^-)$. We encourage the model to assign a higher likelihood to the reference reasoning paths, which is achieved by a standard language modeling loss~\cite{bengio2003neural} to the reference reasoning path $\tau^*$, conditioned on the input question $Q$:
\begin{equation}
\label{eq:ref}
\mathcal{J}_{\text{Ref}}(\theta) = - \mathbb{E}_{(v_{i-1}, v_i) \in \tau^*} [\log p_\theta(v_i | v_{i-1})],
\end{equation}
where $p_\theta(v_i | v_{i-1})$ is the conditional probability of transitioning from state $v_{i-1}$ to $v_i$, and $Q \in v_0$.

Regarding the contrastive reasoning path pairs, since their comparison reveals the correct model optimization direction, we define an alignment loss that provides contrastive feedback between the favorable and unfavorable branches, with the goal of maximizing the likelihood gap between the positive and negative instances. To be more specific, this loss is defined via a log-ratio preference functional~\cite{hong2024orpo}. Let $\mathcal{L}(\tau | \theta) = \log p_\theta(\tau | Q)$ be the log-likelihood of a complete reasoning path. The alignment loss at state $v_i$ is given by:
\begin{equation}
\mathcal{J}_{\text{Align}, i}(\theta) = - k \cdot \log\sigma \left( \mathcal{L}(\tau_i^+ | \theta) - \mathcal{L}(\tau_i^- | \theta) \right),
\end{equation}
where k is a hyperparameter that acts as a scaling factor to control the strength of this alignment loss. 
Notably, since $\tau_i^+$ and $\tau_i^-$ share the same previous state $\tau_{1:i-1}$, their log-likelihood difference simplifies to the difference between the log-likelihoods of their continuation parts:
\begin{equation}
\begin{gathered}
\mathcal{L}(\tau_i^+ | \theta) - \mathcal{L}(\tau_i^- | \theta) 
= \log \frac{p_\theta(\tau'^+_{\text{cont}} | \tau_{1:i-1})}{p_\theta(\tau'^-_{\text{cont}} | \tau_{1:i-1})}.
\end{gathered}
\end{equation}

The total alignment loss is the sum of the losses over all intermediate states:
\begin{equation}
\label{eq:Align}
\mathcal{J}_{\text{Align}}(\theta) = \sum_{i=1}^{T^*-1} \mathcal{J}_{\text{Align}, i}(\theta).
\end{equation}

Finally, the total loss in our framework is a linear combination of Equations (\ref{eq:ref}) and (\ref{eq:Align}):
\begin{equation}
\min_{\theta} \mathcal{J}(\theta) = \mathcal{J}_{\text{Ref}}(\theta) + \lambda \cdot \mathcal{J}_{\text{Align}}(\theta),
\end{equation}
where $\lambda$ is a hyperparameter weight balancing the optimization of the reference reasoning path against that of the contrastive reasoning paths.

\section{Experiments}

\subsubsection{Experiment Setup} To demonstrate the applicability of the proposed SRPO framework, we conduct evaluations on two MLLMs: \texttt{LLaVA-NeXT-LLaMA3}~\cite{liu2024llavanext} and \texttt{Qwen2.5-VL-7B}~\cite{bai2025qwen2}. These models are strategically selected from distinct and influential architectural families to facilitate a rigorous validation of our method's effectiveness across diverse foundations within a focused experimental scope.


\subsubsection{SRPO Implementation} To train the SRPO framework, we employ LoRA fine-tuning~\cite{hu2022lora}, with a fixed batch size of 8, a learning rate of 5e-5, and a LoRA rank of 8. The loss weight $\lambda$ is set to 0.3, which is based on the improvement of our \texttt{Qwen2.5-SRPO} model on the proposed \textit{Reasoning Path Bechmark} (RSBench) with the candidate values {0.1, 0.3, 0.5, 0.7, 0.9}; this value of $\lambda$ is applied to all benchmarks. We also adopt a fixed temperature parameter of 0.5 to sample multiple outputs from the model. All training procedures are conducted on 8 $\times$ A100 GPUs. More training details can be found in Appendix A.4.

\begin{table*}[t]
\centering
\small 
\setlength{\tabcolsep}{4pt} 
\begin{tabular}{l@{\hspace{3pt}}c@{\hspace{6pt}}c@{\hspace{6pt}}c@{\hspace{6pt}}c@{\hspace{6pt}}c@{\hspace{8pt}}c@{\hspace{6pt}}c@{\hspace{6pt}}c@{\hspace{6pt}}c@{\hspace{6pt}}c@{\hspace{6pt}}c}
\toprule
\multirow{2}{*}{\textbf{Models}} & \multicolumn{3}{c}{\textbf{USBBench}} & \multicolumn{2}{c}{\textbf{MLLMGuard}} & \textbf{MSSBench} & \multicolumn{2}{c}{\textbf{SafeBench}} & \textbf{VLSBench} & \multirow{2}{*}{\textbf{Average↑}} & \multirow{2}{*}{\textbf{Average↓}} \\
\cmidrule(lr){2-4} \cmidrule(lr){5-6} \cmidrule(lr){7-7} \cmidrule(lr){8-9} \cmidrule(lr){10-10} 
& ASR↓ & ARR↓ & Avg(\%)↓ & PAR↑ & ASD↓ & Avg(\%)↑ & ASR↓ & SRI↑ & Avg(\%)↑ & & \\
\midrule
\multicolumn{12}{c}{\textit{Closed-source MLLMs}} \\
\midrule
Claude3.5-Sonnet2 & \textbf{32.80} & 25.79 & 29.30 & 52.38 & 9.01 & 69.01 & \textbf{0.70} & \textbf{99.30} & \textbf{79.35} & 75.01 & 13.00 \\
Gemini-1.5-Pro & 64.45 & 11.33 & 37.89 & 38.12 & 21.94 & 61.94 & 2.60 & 97.10 & 50.21 & 61.84 & 20.80 \\
Gemini-2.0-Flash & 76.55 & 5.43 & 40.99 & 45.32 & 20.52 & 65.52 & 2.80 & 97.30 & 52.48 & 65.16 & 21.44 \\
GPT-4o & 72.83 & 3.77 & 38.30 & 56.68 & 14.32 & 59.30 & 3.40 & 96.10 & 69.50 & 70.40 & 18.67 \\
\midrule
\multicolumn{12}{c}{\textit{Open-source MLLMs}} \\
\midrule
DeepSeek-VL & 82.12 & 7.78 & 44.95 & 25.32 & 35.33 & 50.40 & 33.10 & 75.20 & 20.35 & 42.82 & 37.79 \\
VILA-1.5-7B & 88.68 & 32.15 & 60.42 & 16.32 & \textbf{7.65} & 52.23 & 42.30 & 69.80 & 13.56 & 37.98 & 36.79 \\
MiniGPT-v2 & 89.12 & 12.30 & 50.71 & 49.70 & 27.01 & 50.60 & 38.80 & 71.50 & 20.35 & 48.04 & 38.84 \\
LLaVA-v1.5-7B & 84.51 & 8.56 & 46.53 & 20.63 & 43.08 & 56.80 & 39.60 & 72.30 & 8.65 & 39.60 & 43.07 \\
LLaVA-v1.6-mistral-7B & 82.28 & 10.26 & 46.27 & 23.25 & 43.58 & 57.25 & 32.50 & 72.80 & 15.32 & 42.16 & 40.78 \\
InternVL2.5-8B & 80.77 & 11.98 & 46.38 & 40.19 & 32.40 & 51.22 & 21.90 & 82.10 & 21.37 & 48.72 & 33.56 \\
MiniCPM-LLaMA3-V 2.5 & 78.85 & 6.12 & 42.49 & 26.81 & 31.12 & 48.25 & 30.50 & 74.90 & 17.60 & 41.89 & 34.70 \\
MiniCPM-V-2.6 & 81.34 & 6.42 & 43.88 & 32.23 & 33.43 & 47.38 & 28.70 & 75.50 & 15.98 & 42.77 & 35.34 \\
Qwen2-VL-7B & 80.99 & 6.27 & 43.63 & 35.72 & 28.36 & 53.20 & 35.40 & 72.30 & 15.77 & 44.25 & 35.80 \\
GLM-4v-9B & 77.72 & 5.95 & 41.84 & 23.41 & 45.30 & 50.85 & 12.20 & 89.30 & 22.64 & 46.55 & 29.78 \\
\midrule
LLaVA-NeXT-LLaMA3 & 78.88 & 7.53 & 43.20 & 26.45& 42.27 & 56.35 & 29.40 & 73.10 & 18.56 & 41.12 & 38.29 \\
\rowcolor{gray!15}
\quad + SRPO & 50.36 & 6.20 & 28.28 & 50.22 & 17.63 & 71.20 & 7.20 & 91.80 & 50.74 & 65.99 & 14.37 \\
\midrule
Qwen-2.5VL-7B & 75.26 & 4.72 & 39.99 & 38.22 & 28.35 & 55.36 & 32.20 & 76.30 & 20.45 & 47.58 & 33.51 \\
\rowcolor{gray!15}
\quad + SRPO & 42.38 & \textbf{2.83} & \textbf{22.60} & \textbf{65.30} & 7.92 & \textbf{73.89} & 7.50 & 98.10 & 67.43 & \textbf{76.18} & \textbf{12.67} \\
\bottomrule
\end{tabular}
\caption{Safety evaluation results on 5 benchmarks. Applying our SRPO framework significantly promotes the safety performance of both LLaVA-NeXT-LLaMA3 and Qwen2.5-VL, facilitating them to surpass other state-of-the-art MLLMs. 
}
\label{tab:visual_reasoning_results}
\end{table*}

\subsubsection{Evaluated Models and Configurations}
We evaluate both open-source and closed-source MLLMs. For open-source MLLMs, recently released mainstream models are taken into consideration, which include \texttt{Qwen2.5-VL} series~\cite{bai2025qwen2}, \texttt{Qwen2-VL} series~\cite{wang2024qwen2}, \texttt{InternVL2} series~\cite{chen2024internvl}, \texttt{GLM-4V}~\cite{glm2024chatglm}, \texttt{LLaVA-v1.5} series~\cite{liu2024improved}, \texttt{DeepSeek-VL}~\cite{lu2024deepseek}, \texttt{MiniGPT-v2}~\cite{chen2023minigpt}, \texttt{MiniCPM-v2.6}~\cite{yao2024minicpm}, and \texttt{VILA} series~\cite{lin2024vila}. For close-source commercial MLLMs, we select \texttt{GPT-4o}, \texttt{Claude-3.5-Sonnet2}, and the \texttt{Gemini} series. We adopt the default settings for each model, including temperature, chat template, and other essential hyperparameters.

\subsubsection{Benchmark Setup}
Our experiments are conducted on various multimodal safety benchmarks. For example, we adopt USBBench~\cite{zheng2025usb} and MSSBench~\cite{zhou2024multimodal} for contextual safety, with a specific focus on the more challenging SIST subset of USBBench. Furthermore, we adopt MLLM-GUARD~\cite{gu2024mllmguard}, a multi-dimensional safety suite assessing five key safety dimensions; SafeBench~\cite{ying2024safebench}, a comprehensive framework that evaluates MLLMs against a detailed taxonomy of 8 primary risk categories and 23 sub-categories; and VLSBench~\cite{hu2024vlsbench}, a reliable cross-modal benchmark structured around a safety taxonomy of 6 main categories and 19 sub-categories.

\begin{figure}[t]
\centering
\includegraphics[width=\columnwidth]{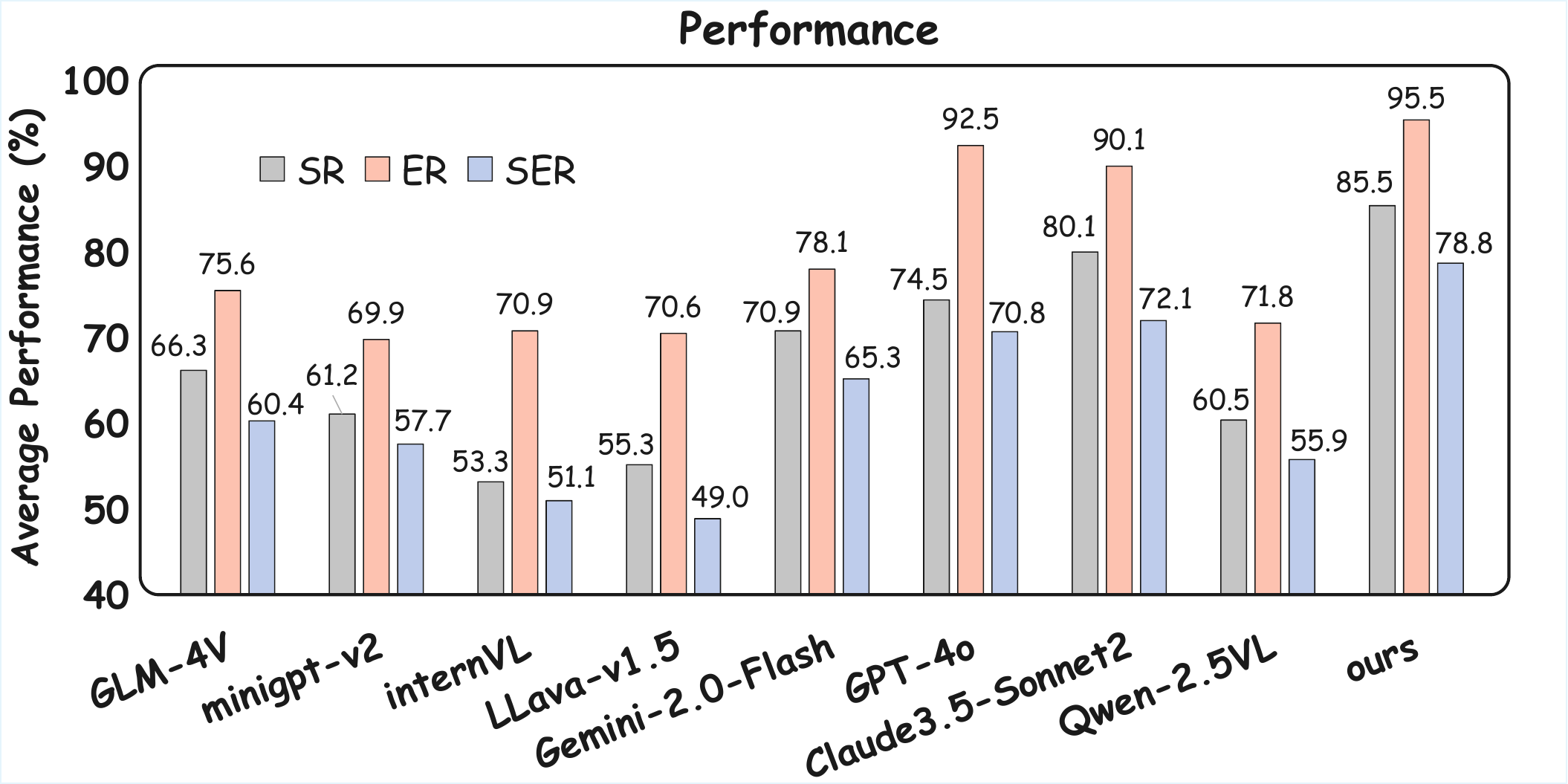} 
\caption{Main results on the proposed RSBench. Our SRPO framework outperforms other methods in both safety and effectiveness of the reasoning paths.}
\label{fig5}
\end{figure}

\subsection{Main Results}
The experimental results in Table~\ref{tab:visual_reasoning_results} demonstrate the effectiveness and generalizability of our proposed SRPO framework in enhancing the safety capabilities of MLLMs. By applying SRPO to \texttt{LLaVA-NeXT-LLaMA3} and \texttt{Qwen2.5-VL}, both models achieve substantial gain on the selected challenging cross-modal safety benchmarks. On average, \texttt{LLaVA-NeXT-LLaMA3} and \texttt{Qwen2.5-VL} exhibit performance improvements of 24.87\% and 28.60\%, respectively, accompanied by reductions in the attack success rate (ASR) of 23.92\% and 20.84\%. These findings underscore the effectiveness of SRPO in strengthening safety reasoning among various MLLMs.

Specifically, on MSSBench and VLSBench, \texttt{Qwen2.5-SRPO} achieves notable improvements of 18.53\% and 46.98\%, respectively. The remarkable gain on VLSBench highlights the enhanced capability of SRPO in handling more challenging and nuanced safety risks. On USBBench, MLLM-GUARD, and SafeBench, \texttt{Qwen2.5-SRPO} reduced the ASR by 17.39\%, 20.43\%, and 24.7\%, respectively. In addition, the model’s safety-issues-detection ability improved by 27.08\% on MLLM-GUARD and 21.8\% on SafeBench. Following the integration of SRPO, both \texttt{LLaVA-NeXT-LLaMA3} and \texttt{Qwen2.5-VL} exhibit strong safety performance that exceeds most commercial MLLMs. The results above further validate the effectiveness of our method in enhancing the safety reasoning capabilities of MLLMs.


\subsection{RSBench}
While existing evaluations predominantly focus on assessing the final output of the model, they often overlook the quality of the intermediate CoTs of the safety reasoning process. To address this issue, we introduce the \textit{Reasoning Path Benchmark} (RSBench), which provides a more comprehensive evaluation of  MLLM's safety reasoning capabilities. RSBench leverages \texttt{GPT-4o} as an arbitration model and introduces two key metrics: \textit{safety rate} (SR) and \textit{effectiveness rate} (ER). SR quantifies the proportion of reasoning paths deemed safe, while the ER captures the proportion of reasoning paths considered practically useful. Formally, these metrics are defined as:
\begin{equation}
SR = \frac{1}{N} \sum_{i=1}^{N_h} f_h(i) \ \mbox{,} \ ER = \frac{1}{N}\sum_{j=1}^{N_r} f_r(j),
\end{equation}
where $N_h$, $N_r$, and $N$ represent the number of safe responses, effective responses, and total responses, respectively. $f_h(i)$ and $f_r(j)$ are indicator functions. $f_h(i)$ equals to 1 if the $i$-th query yields a safe response and 0 otherwise. Similarly, $f_r(j)$ equals 1 if the $j$-th query yields an effective response and 0 otherwise.

To enable a unified evaluation of both safety and effectiveness in the CoT reasoning process, we further define the \textit{safety and effectiveness rate} (SER), which quantifies the proportion of reasoning paths that simultaneously satisfy both safety and effectiveness criteria:
\begin{equation}
SER = \frac{1}{N} \sum_{k=1}^{N} [f_h(k) \cdot f_r(k)].
\end{equation}

As shown in Figure~\ref{fig5}, we evaluate our proposed model \texttt{Qwen2.5-SRPO} and 8 families of advanced MLLMs on RSBench. The experimental results indicate that \texttt{Qwen2.5-SRPO} significantly outperforms its base model \texttt{Qwen-2.5VL}, with a more than 20\% absolute gain in both SR and ER. Furthermore, \texttt{Qwen2.5-SRPO} also exhibits safer and more effective reasoning paths against the selected leading closed-source MLLMs.

\begin{figure}[t]
\centering
\includegraphics[width=0.8\columnwidth]{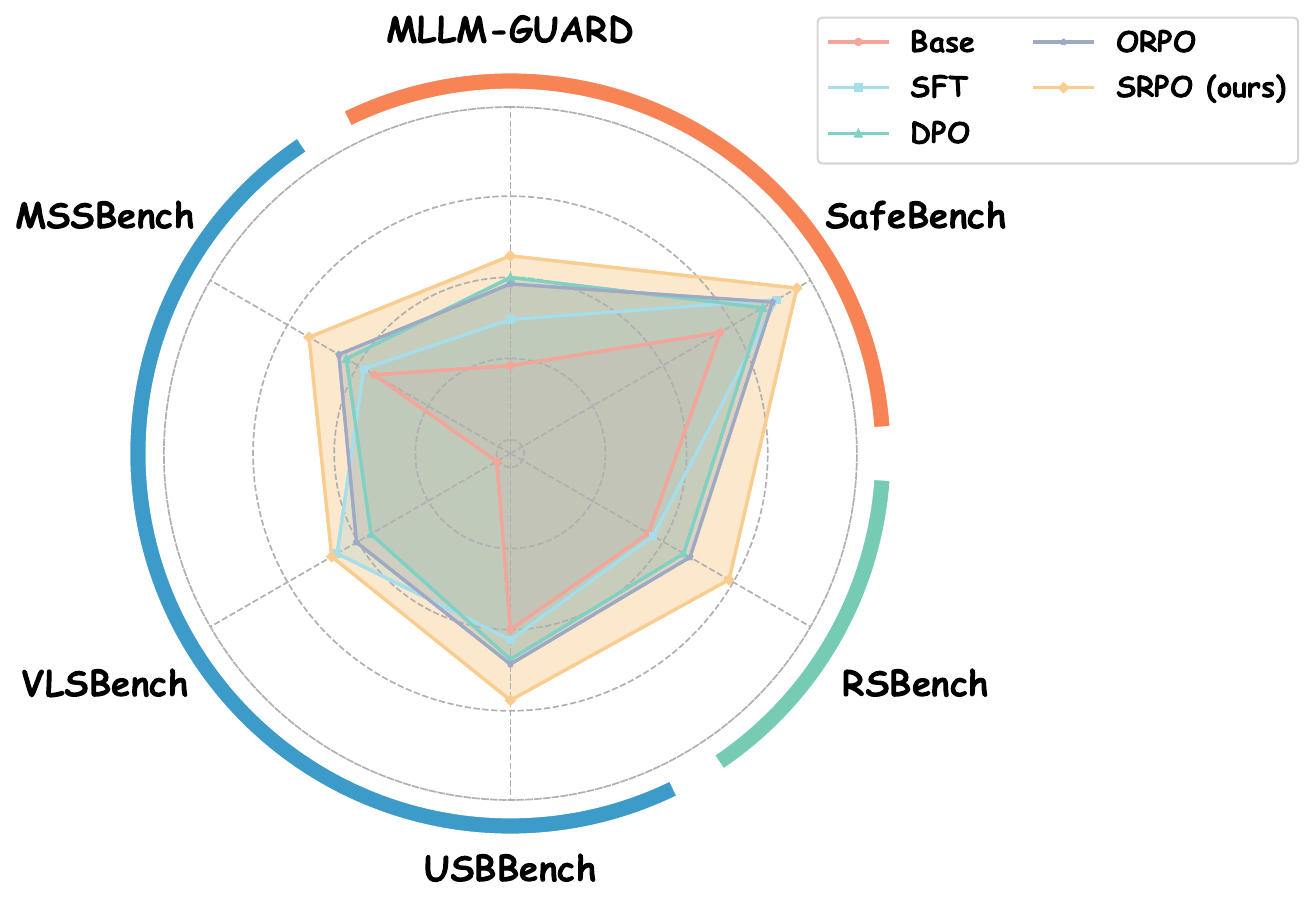} 
\caption{Main results of our proposed SRPO framework against different optimization baselines on six benchmarks. 
}
\label{fig6}
\end{figure}

\subsection{Optimization Baselines}

To further validate the effectiveness of our proposed method, we conduct a comparative analysis against several popular optimization baselines, which include both reasoning-focused training approaches and preference-based optimization techniques, \textit{i.e.} SFT, DPO and ORPO. With \texttt{Qwen-2.5VL} as the base model, we apply each optimization method and evaluate their performance alongside \texttt{Qwen2.5-SRPO} on the five previously introduced benchmarks as well as our proposed RSBench.

\begin{figure}[t]
\centering
\includegraphics[width=\columnwidth]{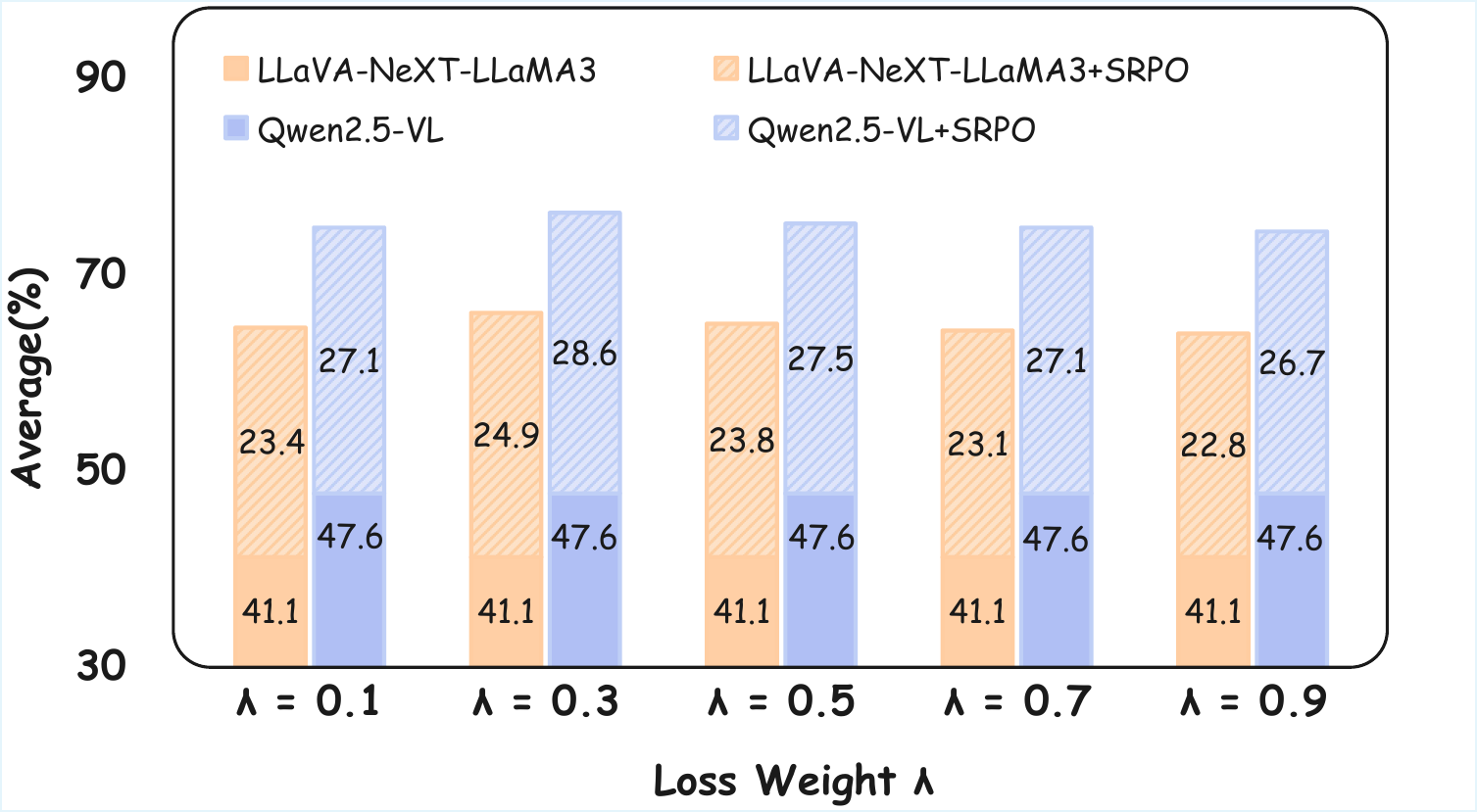}
\caption{Influence of the loss weight $\lambda$ on the safety alignment performance of MLLMs.}
\label{fig7}
\end{figure}

To ensure a fair comparison, we standardized the data setup so that each optimization baseline is trained on all applicable samples. Specifically, SFT utilizes the entire dataset, as its format requires only the input question and the ground-truth answer. DPO and ORPO, on the other hand, are exclusively trained on samples from our SSUI dataset that include at least one correct and one incorrect reasoning path generated from the exploration stage. All optimization baselines except SFT employ a fixed temperature parameter when generating reasoning paths via CoT prompting.

As demonstrated in Figure~\ref{fig6}, the results reveal that our method consistently outperforms all optimization baselines among the evaluated benchmarks. The performance gains are particularly visible on the more challenging datasets such as MSSBench, USBBench, and the CoT-centric RSBench, highlighting our method’s superior ability to learn from the explored reasoning paths. In contrast, SFT generally underperforms in approaches leveraging self-explored reasoning, particularly on the more challenging benchmarks. This suggests that while directly predicting the correct answer may suffice in simpler cases, it is less effective for tasks demanding nuanced safety reasoning.

\subsection{Further Analysis}
To investigate the effect of reasoning exploration within our framework, an analysis of the loss weight parameter $\lambda$ is conducted. Specifically, a smaller $\lambda$ emphasizes more of the reference path to a safe answer. Conversely, a larger $\lambda$ assigns more importance to the favorable and unfavorable branches generated at each reasoning step. As illustrated in Figure~\ref{fig7}, an excessively small $\lambda$ yields suboptimal results, as it inadequately emphasizes reasoning exploration. Similarly, over-emphasizing exploration is not beneficial for training either, since sufficient grounding of the model in the reference path remains crucial. As a result, a trade-off between optimizing for the reference reasoning path and the exploratory branches is required.

\section{Conclusion}
%
In this work, we address the critical challenge of \textit{implicit reasoning risk} in MLLMs by introducing the Safety-Aware Reasoning Path Optimization framework. Supported by our proposed SSUI dataset and RSBench benchmark, SRPO leverages generative exploration and contrastive optimization to steer the model towards safe reasoning paths. 
Extensive experiments demonstrate that our SRPO-enhanced model achieves SOTA results on key safety benchmarks, outperforming even leading commercial MLLMs.
These results verify that aligning the reasoning process itself is a more robust safety strategy than merely filtering outputs, thereby presenting a new paradigm for building fundamentally more trustworthy AI by ensuring the integrity of their thought processes.


\bibliography{aaai2026}

\end{document}